\documentclass[runningheads]{llncs}
\usepackage[english]{babel}
\usepackage{makecell}
\usepackage{graphicx}
\usepackage{amssymb}
\usepackage{amsmath}
\usepackage[table]{xcolor}
\usepackage{multirow}
\usepackage{tikz}
\usepackage{hyperref}
\usepackage{cite}
\usepackage{mymacros}
\usepackage{tikz}
\usepackage{paralist}
\usepackage{color,soul}

\begin{document}
\title{
Monitoring Constraints in Business Processes Using Object-Centric Constraint Graphs
% A Log-based Approach to Monitoring Object-Centric Constraints in Business Processes
% :\\ An Object-Centric Approach
}
\titlerunning{Monitoring Object-Centric Constraints in Business Processes}
\author{
Gyunam Park
% \orcidID{0000-0001-9394-6513} 
\and
Wil. M. P. van der Aalst
% \orcidID{0000-0002-0955-6940}
}
\authorrunning{G. Park and W.M.P. van der Aalst}
\institute{Process and Data Science Group (PADS), RWTH Aachen University \\ \email{\{gnpark,wvdaalst\}@pads.rwth-aachen.de}}
\maketitle              % typeset the header of the contribution
\begin{abstract}
Constraint monitoring aims to monitor the violation of constraints in business processes, e.g., an invoice should be cleared within 48 hours after the corresponding goods receipt, by analyzing event data.
% Based on the insights, an organization can redesign the process or enforce best practices.
Existing techniques for constraint monitoring assume that a single case notion exists in a business process, e.g., a patient in a healthcare process, and each event is associated with the case notion.
However, in reality, business processes are \emph{object-centric}, i.e., multiple case notions (objects) exist, and an event may be associated with multiple objects.
For instance, an Order-To-Cash (O2C) process involves \textit{order}, \textit{item}, \textit{delivery}, etc., and they interact when executing an event, e.g., packing multiple items together for a delivery.
The existing techniques produce misleading insights when applied to such object-centric business processes.
In this work, we propose an approach to monitoring constraints in object-centric business processes.
To this end, we introduce \emph{Object-Centric Constraint Graphs} (OCCGs) to represent constraints that consider the interaction of objects.
Next, we evaluate the constraints represented by OCCGs by analyzing Object-Centric Event Logs (OCELs) that store the interaction of different objects in events.
We have implemented a web application to support the proposed approach and conducted two case studies using a real-life SAP ERP system.

% It facilitates process analysts to 1) elicit operational problems, enabling the monitoring of such problems, and 2) define management actions to deal with the problems, enabling the automated execution of such actions.
% Recently, Object-Centric Petri Nets (OCPNs) were introduced to realistically represent operational processes by considering \emph{concurrency}, referring that different parts of an organization operate concurrently, and \emph{object-centricity}, referring that multiple objects (i.e., multiple case notions) exist and they interact in the process.
% In this work, we aim to support the elicitation, representation, and monitoring of problem patterns existing in organizations' operational processes using OCPNs.
% To this end, we 1) provide common problem patterns expressed in OCPNs to support the elicitation of problem patterns, 2) develop pattern graphs to formally represent the patterns, and 3) suggest a data-driven approach to monitoring the patterns.
% We have implemented a web application to support the proposed approach and conducted two case studies using a real-life SAP ERP system.
% \keywords{Process-Centric Problem Patterns \and Common Problem Patterns \and Pattern Graphs \and Constraint Monitoring \and Process Mining}
\end{abstract}
\section{Introduction}
It is indispensable for organizations to continuously monitor their operational problems and take proactive actions to mitigate risks and improve performances~\cite{DBLP:books/sp/Aalst16}.
Constraint monitoring aims at detecting violations of \emph{constraints} (i.e., operational problems) in business processes of an organization by analyzing event data recorded by information systems~\cite{DBLP:journals/is/LyMMRA15}.
Once violations are detected, the organization can redesign its process to cover the respective violation.

A plethora of techniques has been suggested to implement constraint monitoring.
For instance, in \cite{DBLP:conf/bpm/WeidlichZMGWD11}, a technique is proposed to detect events violating constraints, e.g., detecting an X-ray event with a long waiting time, using behavioral profiles and Complex Event Processing (CEP).
In \cite{DBLP:conf/rv/MaggiWMA11}, authors propose a technique to detect process instances violating constraints, e.g., detecting a patient with multiple executions of X-rays, using Linear Temporal Logic (LTL).
% They are mainly classified into three categories: \emph{event-level}, \emph{case-level}, and \emph{process-level} constraint monitoring.
% Techniques for event-level constraint monitoring aim at detecting events violating constraints (e.g., detecting an X-ray event with a long waiting time)~\cite{DBLP:conf/bpm/WeidlichZMGWD11}, whereas ones for case-level constraint monitoring aim to monitor the violations of constraints by process instances (e.g., detecting a patient with multiple executions of X-rays)~\cite{DBLP:conf/rv/MaggiWMA11}.
% Finally, process-level constraint monitoring evaluates if a (part of) process violates constraints defined over the process (e.g., detecting the long average waiting time for X-ray)~\cite{DBLP:conf/bpm/AwadDW08}.

The existing techniques assume that an event in event data is associated with a single object of a unique type (so-called case), e.g., a patient in a healthcare process.
Thus, constraints are defined over the single case notion, e.g., each patient (i.e., case) should be registered before triage.
However, in real-life business processes, an event may be associated with multiple objects of different types, i.e., real-life business processes are \emph{object-centric}~\cite{DBLP:conf/sefm/Aalst19}.
% We call such business processes \emph{object-centric} processes.
For instance, the omnipresent Purchase-To-Pay (P2P) process involves different object types, e.g., \textit{purchase order}, \textit{material}, \textit{invoice}, \textit{goods receipt}, etc., and an event may be associated with multiple objects of different types, e.g., clearing invoice is associated with a purchase order, an invoice, and a goods receipt to enable so-called \textit{three-way matching}.

\begin{figure}[!htb]
    \centering
    \includegraphics[width=1\linewidth]{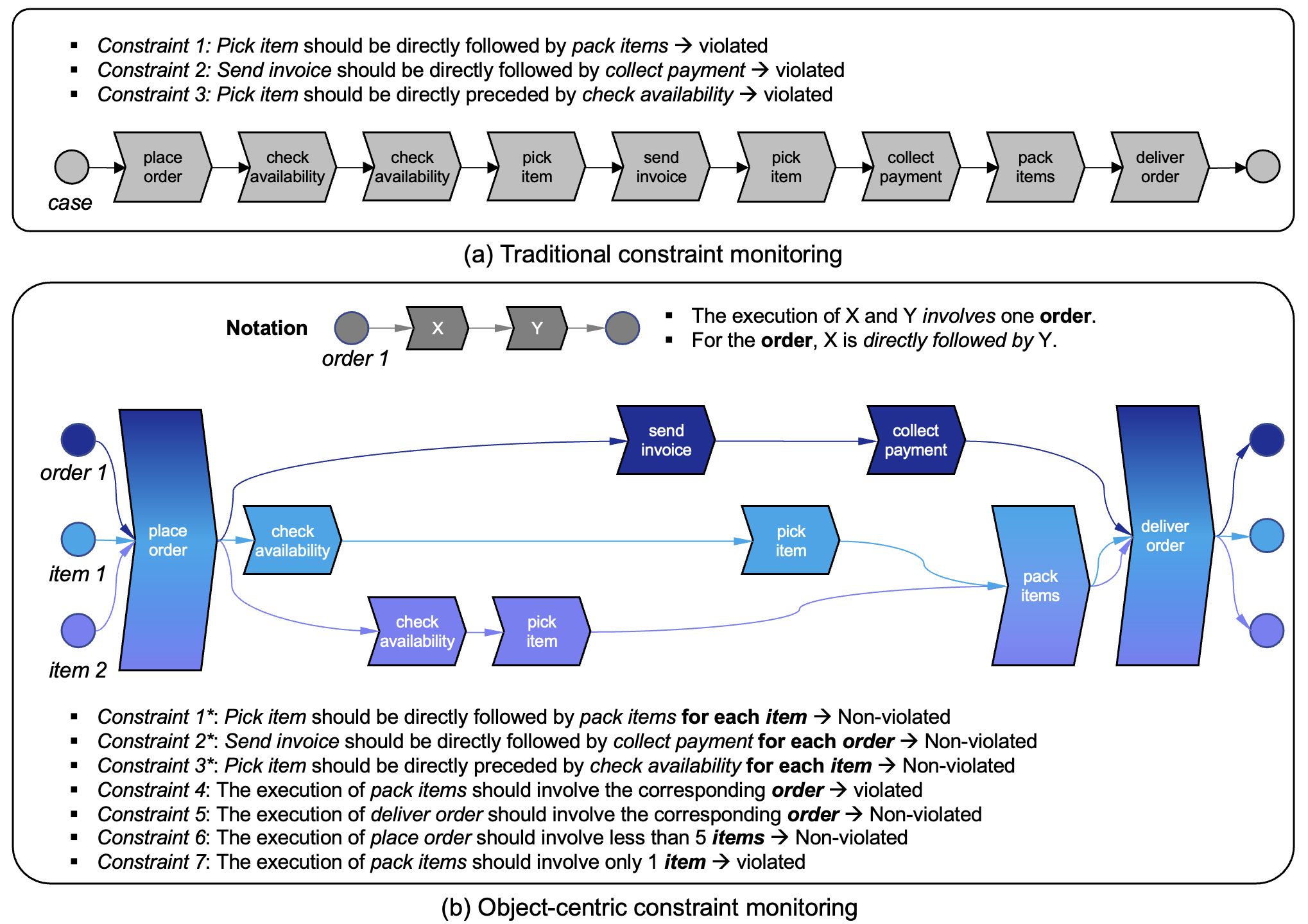}
    \caption{Comparing (a) traditional and (b) object-centric constraint monitoring}
    \label{fig:intro-example}
\end{figure}

Applying the existing techniques to such object-centric settings results in misleading insights.
\autoref{fig:intro-example}(a) shows events of a ``case'' in an Order-To-Cash (O2C) process using \textit{order} as the case notion.
First, an order is placed, and the availability of two items of the order is checked, respectively.
Next, one of the items is picked, and the invoice of the order is sent to the customer.
Afterward, the other item is picked, and the payment of the invoice is collected.
Finally, the items are packed and delivered to the customer.
The three constraints shown in \autoref{fig:intro-example}(a) are violated by the case.
For instance, \textit{Constraint 1} is violated since \textit{pick item} is followed by \textit{send invoice} in the case and \textit{Constraint 3} is violated since \textit{pick item} is preceded by \textit{send invoice}.

However, in reality, the order and each item have different lifecycles as shown in \autoref{fig:intro-example}(b).
First, we place an order with two items.
While the invoice is sent and the payment is collected for the order, we check the availability of each item and pick each of them.
We finally deliver the order with two items after packing two items together.
In this object-centric setting, constraints should be defined in relation to objects to provide accurate insights.
For instance, \textit{Constraint 1*} extends \textit{Constraint 1} with the corresponding object type (i.e., item).
Contrary to \textit{Constraint 1}, \textit{Constraint 1*} is not violated since \textit{pick item} is directly followed by \textit{pack item} for any items.
% The constraints in \autoref{fig:intro-example}(a) should be defined over each object type, e.g., .
% In fact, the three constraints in \autoref{fig:intro-example}(b) are not violated, e.g., pick item is always directly followed by pack items in the item's lifecycle.
Moreover, we can analyze more object-centric constraints by considering the interaction of different objects.
First, we can analyze if an execution of an activity involves (un)necessary objects (cf. \textit{Constraint 4} and \textit{Constraint 5}).
Also, we can analyze the cardinality of objects for executing an activity (cf. \textit{Constraint 6} and \textit{Constraint 7}).

In this work, we propose a technique for constraint monitoring in object-centric settings.
To this end, we first introduce \emph{object-centric behavioral metrics} that can be computed from Object-Centric Event Logs (OCELs), e.g., a metric to measure the degree to which \textit{pick item} precedes \textit{pack items} in the lifecycle of \textit{items}.
% ($metric_1$).
% Such metrics are computed using OCELs.
Next, we develop Object-Centric Constraint Graphs (OCCGs) to formally represent constraints using such metrics.
% , e.g., $metric_1$ should be less than $0.6$ ($const_1$).
Finally, \emph{monitoring engine} evaluates the violation of the constraints represented by OCCGs by 
analyzing OCELs.
% computing the metrics using an OCEL, e.g., $metric_1$ is computed to be $0.9$ from the event log, and thus, $const_1$ is violated.

We have implemented a web application to support the approach. 
A demo video and a manual are available at \url{https://github.com/gyunamister/ProPPa.git}.
Moreover, we have conducted case studies with a production process and a Purchase-To-Pay (P2P) process supported by an SAP ERP system.

The remainder is organized as follows.
We discuss the related work in \autoref{sec:related} and present the preliminaries, including OCELs in \autoref{sec:preliminaries}.
In \autoref{sec:behavioral-metrics}, we introduce object-centric behavioral metrics.
Afterward, we present OCCGs to formally represent constraints and the monitoring engine to evaluate the violation of the constraints in \autoref{sec:pattern-graphs}.
Next, \autoref{sec:evaluation} introduces the implementation of the proposed approach and case studies using real-life event data.
Finally, \autoref{sec:conclusion} concludes the paper.
\section{Related Work}\label{sec:related}
Many approaches have been proposed to monitor the violation of constraints by analyzing event data.
% to 1) formally represent problems existing in business processes to certain formats of constraints and 2) monitor the violation of constraints using the representation.
% Constraint monitoring techniques are classified into three categories: \emph{event-level}, \emph{case-level}, and \emph{process-level} constraint monitoring.
% First, event-level constraint monitoring techniques aim to detect violated executions of events.
Weidlich et al.~\cite{DBLP:conf/bpm/WeidlichZMGWD11} propose a technique to abstract process models to behavioral profiles and produce event queries from the profile.
Violated executions of events are monitored using Complex Event Processing (CEP) engines with the event queries.
Awad et al.~\cite{DBLP:conf/sac/AwadBESAS15} define a set of generic patterns regarding the occurrence of tasks, their ordering, and resource assignments and generate anti-patterns from the generic patterns to monitor event executions.
% Second, case-level constraint monitoring techniques evaluate constraints defined over process instances, aiming to detect process instances violating the constraints.
Maggi et al.~\cite{DBLP:conf/rv/MaggiWMA11} represent control-flow properties of a running process instance using Linear Temporal Logic (LTL) and evaluate their violations at runtime.
% A global automaton, representing the conjunction of all constraints, is used to identify possible conflicts among different constraints.
% Alignments are often used to check the conformance of process instances~\cite{DBLP:journals/widm/AalstAD12}.
% Van Zelst et al.~\cite{DBLP:journals/ijdsa/ZelstBHDA19} suggests a technique to compute prefix-alignments to detect deviant behavior.
In~\cite{DBLP:conf/bpm/RamezaniFA12}, Petri-net-based constraints are aligned with event logs to evaluate whether the execution of business processes conforms to the constraints.
The existing techniques may produce misleading insights in object-centric settings since it does not consider the interaction among objects.
Moreover, object-centric constraints, e.g., the cardinality of an object type for the execution of an activity, are not supported in the existing techniques.

% This paper focuses on process-level constraint monitoring, where the goal is to evaluate if a process violates constraints defined over the process.
% % Most of the existing literature can be found in the research field of process querying~\cite{DBLP:journals/dss/PolyvyanyyOBA17}.
% Awad et al.~\cite{DBLP:conf/bpm/AwadDW08} represent a compliance rule as a BPMN-Q
% % , translate it to temporal logic formula, 
% and evaluate if a process model satisfies the compliance rule.
% In~\cite{DBLP:conf/apbpm/HofstedeORS0P13}, A Process-Model Query Language (APQL) is proposed to retrieve process models from process model repositories based on semantic relationships between activities in process models.
% % The language uses a variety of basic temporal relationships between activities, e.g., precedence and succession, supporting the composition of more complex relationships using the basic relationships.
% Jin et al.~\cite{DBLP:conf/dasfaa/JinWW11} propose a text-based language, called Behavior Query Language (BQL), to query Petri nets that contain certain control-flow behaviors based on the semantics of Petri nets.

This paper is in line with the recent developments in object-centric process mining~\cite{DBLP:conf/sefm/Aalst19}.
Object-centric process mining breaks the assumption of traditional process mining techniques that each event is associated with a single case notion (i.e., object), allowing one event to be associated with multiple objects.
% An event log format has been proposed to store Object-Centric Event Logs (OCELs)~\cite{DBLP:conf/adbis/GhahfarokhiPBA21}.
In \cite{DBLP:journals/fuin/AalstB20}, a process discovery technique is proposed to discover Object-Centric Petri Nets (OCPNs) from OCELs.
A conformance checking technique to determine the precision and fitness of the net is suggested in \cite{DBLP:conf/icpm/AdamsA21}, and an approach to object-centric performance analysis is proposed in 
\cite{https://doi.org/10.48550/arxiv.2204.10662}.
Esser and Fahland~\cite{DBLP:journals/jodsn/EsserF21} propose a graph database as a storage format for object-centric event data, enabling a user to use queries to calculate different statistics.
This work extends the current development in the field of object-centric process mining by proposing a constraint monitoring technique in object-centric settings.
% lacks the consideration of object-centricity, i.e., the language does not support querying processes based on the interaction of different objects in the execution of activities, only focusing on the existence/ordering of such activities.
\section{Preliminaries}\label{sec:preliminaries}
Given a set $X$, the powerset $\pow(X)$ denotes the set of all possible subsets.
We denote a sequence with $\sigma = \langle x_1, \ldots , x_n \rangle$ and the set of all sequences over $X$ with $X^*$. 
Given a sequence $\sigma \in X^*$, $x \in \sigma$ if and only if $\exists_{1 \le i \le |\sigma|} \; \sigma(i)=x$.

\begin{definition}[Universes]\label{def:universes}
$\univ{ei}$ is the universe of event identifiers, 
$\univ{oi}$ is the universe of object identifiers,
$\univ{act}$ is the universe of activity names, 
$\univ{time}$ is the universe of timestamps, 
$\univ{ot}$ is the universe of object types,
$\univ{attr}$ is the universe of attributes,
$\univ{val}$ is the universe of values, and
$\univ{map} = \univ{attr} \nrightarrow \univ{val}$ is the universe of attribute-value mappings.
For any $f \in \univ{map}$ and $x \notin \mi{dom}(f)$, $f(x)=\perp$.
\end{definition}

Using the universes, we define an object-centric event log as follows.

\begin{definition}[Object-Centric Event Log]\label{def:event}
An object-centric event log is a tuple $L=(E,O,\mu,R)$, where $E \subseteq \univ{event}$ is a set of events, $O \subseteq \univ{oi}$ is a set of objects, $\mu \in (E \rightarrow \univ{map}) \cup (O \rightarrow (\univ{time} \rightarrow \univ{map}))$ is a mapping, and $R \subseteq E \times O$ is a relation, such that for any $e \in E$, $\mu(e)(act) \in \univ{act}$ and $\mu(e)(time) \in \univ{time}$, and for any $o \in O$ and $t,t' \in \univ{time}$, $\mu(o)(t)(\mi{type})=\mu(o)(t')(\mi{type}) \in \univ{ot}$.
$\univ{L}$ is the set of all possible object-centric event logs.
\end{definition}

\begin{table}[t!]
\caption{A fragment of an event log.}\label{tab:ocel}
\centering
\resizebox{0.7\linewidth}{!}{\begin{tabular}{|c|c|c|c|c|}
\hline
  % after \ \hline or \cline{col1-col2} \cline{col3-col4} ...
event id & activity & timestamp & order & item \\ \hline
%  $\ldots$ & $\ldots$ &  $\ldots$ &  $\ldots$ &  $\ldots$\\
$e_{93}$ & \emph{place order (po)} & 25-10-2022:09.35 & $\{o_1\}$ & $\{i_1,i_2,i_3\}$\\
$e_{94}$ & \emph{evaluate credit (ec)} & 25-10-2022:13.35 & $\{o_1\}$ & $\emptyset$\\
$e_{95}$ & \emph{confirm order (co)} & 25-10-2022:15.35 & $\{o_1\}$ & $\{i_1,i_2,i_3\}$\\
% $\ldots$ & $\ldots$ &  $\ldots$ &  $\ldots$ &  $\ldots$\\
\hline
\end{tabular}}
\end{table}

For the sake of brevity, we denote $\mu(e)(x)$ as $\mu_{x}(e)$ and $\mu(o)(t)(x)$ as $\mu_{x}^{t}(o)$. Since the type of an object does not change over time, we denote $\mu_{type}^{t}(o)$ as $\mu_{type}(o)$.
\autoref{tab:ocel} describes a fraction of a simple event log $L_1=(E_1,O_1,\mu_1,R_1)$ with 
$E_1=\{e_{93},e_{94},e_{95}\}$, $O_1=\{o_1,i_1,i_2,i_3\}$, $R_1=\{(e_{93},o_1),(e_{93},i_1),\dots\}$, $\mu_{act}(e_{93})\allowbreak=\textit{po}$, $\mu_{time}(e_{93})=\textit{\scriptsize 25-10-2022:09.35}$, $\mu_{type}(o_1)=\textit{Order}$, and $\mu_{type}(i_1)=\textit{Item}$.

We define functions to query event logs as follows:

\begin{definition}[Notations]\label{def:event}
For an object-centric event log $L=(E,O,\mu,R)$, we introduce the following notations:
\begin{compactitem}
    \item $\mi{acts}(L)=\{\mu_{act}(e) \mid e \in E\}$ is the set of activities,
    \item $\mi{events}(a)=\{e \in E \mid \mu_{act}(e) = a \}$ is the set of the events associated to $a \in \mi{acts}(L)$,
    \item $\mi{types}(L)=\{\mu_{type}(o) \mid o \in O\}$ is the set of object types,
    \item $\mi{objects}(ot)=\{o \in O \mid \mu_{type}(o) = ot \}$ is the set of the objects associated to $ot \in \mi{types}(L)$,
    \item $\mi{events}(o)=\{e \in E \mid (e,o) \in R\}$ is the set of the events containing $o \in O$,
    \item $\mi{objects}(e)=\{o \in O \mid (e,o) \in R\}$ is the set of the objects involved in $e \in E$,
    % \item $objects_{ot}(e)=\{o \in \mi{objects}(e) \mid \mu_{type}(o)=ot\}$ is the set of the objects of $ot \in \mi{types}(L)$ involved in $e \in E$,
    \item $\mi{seq}(o)=\langle e_1,e_2,\dots,e_n \rangle$ such that $\mi{events}(o)=\{e_1,e_2,\dots,e_n\}$ and $\mu_{time}(e_i) \le \mu_{time}(e_j)$ for any $1\le i < j \le n$ is the sequence of all events where object $o\in O$ is involved in, and
    \item $\mi{trace}(o)=\langle a_1,a_2,\dots,a_n \rangle$ such that $\mi{seq}(o)=\langle e_1,e_2,\dots,e_n \rangle$ and $a_i=\mu_{act}(e_i)$ for any $1\le i \le n$ is the trace of object $o \in O$.
\end{compactitem}
\end{definition}

For instance, 
$\mi{acts}(L_1)=\{\textit{po},\textit{ec},\textit{co}\}$,
$\mi{events}(\textit{po})=\{e_{93}\}$,
$\mi{types}(L_1)=\{\textit{Order},\allowbreak \textit{Item}\}$,
$\mi{objects}(\textit{Order})=\{o_1\}$,
$\mi{events}(o_1)=\{e_{93},e_{94},e_{95}\}$, 
% $\mi{events}(i_1)=\{e_{93},e_{95}\}$, 
$\mi{objects}(e_{93})=\{o_1,i_1, \allowbreak i_2,i_3\}$, 
% $\mi{objects}(e_{94})=\{o_1\}$, 
$\mi{seq}(o_1)\allowbreak= \langle e_{93}, e_{94}, e_{95} \rangle$, and
$\mi{trace}(o_1)=\langle \textit{po}, \textit{ec}, \textit{co} \rangle$.

Using the notations, we characterize an event log as follows:

\begin{definition}[Log Characteristics]\label{def:eoopn}
Let $L=(E,O,\mu,R)$ be an object-centric event log.
For $ot \in \mi{types}(L)$ and $a,b\in \mi{acts}(L)$, we define the following characteristics of $L$:
\begin{compactitem}
    % \item $\#_{L}(ot,a)=|\{o \in \mi{objects}(ot) \mid a \in \mi{trace}(o) \}|$ counts the objects of type $ot$ related to activity $a$,
    \item $\#_{L}(ot,X)=|\{o \in \mi{objects}(ot) \mid \forall_{x \in X} \ x \in \mi{trace}(o) \}|$ counts the objects of type $ot$ whose trace contains $X \subseteq \mi{acts}(L)$,
    \item $\#_{L}(ot,a,b)=|\{ o \in \mi{objects}(ot) \mid \exists_{1 \le i < j \le |\mi{trace}(o)|} \ \mi{trace}(o)(i)=a \land \mi{trace}(o)(j)=b \}|$ counts the objects of type $ot$ whose trace contains $a$ followed by $b$,
    \item $\#_{L}^{0}(ot,a)=|\{ e \in \mi{events}(a) \mid |\{o \in \mi{objects}(e) \mid \mu_{type}(o)=ot\}|=0 \}|$ counts the events relating no objects of type $ot$ for the execution of $a$,
    \item $\#_{L}^{1}(ot,a)=|\{ e \in \mi{events}(a) \mid |\{o \in \mi{objects}(e) \mid \mu_{type}(o)=ot\}|=1 \}|$ counts the events relating one object of type $ot$ for the execution of $a$, and
    \item $\#_{L}^{*}(ot,a)=|\{ e \in \mi{events}(a) \mid |\{o \in \mi{objects}(e) \mid \mu_{type}(o)=ot\}|>1 \}|$ counts the events relating more than one object of type $ot$ for the execution of $a$.
    % \item $\mi{skip}_{\mi{AN}}(a)=\mi{true}$ if $\exists_{\sigma_{bv}^1,\sigma_{bv}^2 \in \phi(\mi{AN})} \ a \in \sigma_{bv}^1 \land a \notin \sigma_{bv}^2 $, and $\mi{false}$, otherwise.
\end{compactitem}
\end{definition}

For instance, 
$\#_{L_1}(\textit{Order},\{\textit{po}\})=1$, 
$\#_{L_1}(\textit{Item},\{\textit{po}\})=3$, 
% $\#_{L_1}(\textit{Order},\{\textit{po},\textit{ec}\})=1$, 
$\#_{L_1}(\textit{Item},\{\textit{po},\textit{ec}\})=0$,
$\#_{L_1}(\textit{Order},\textit{po},\textit{ec})=1$,
$\#_{L_1}^{0}(\textit{Order},\textit{ec})=0$, 
$\#_{L_1}^{0}(\textit{Item},\textit{ec})=1$,
$\#_{L_1}^{1}(\textit{Order},\textit{po})=1$,
$\#_{L_1}^{1}\allowbreak(\textit{Item},\textit{po})=0$,
$\#_{L_1}^{*}(\textit{Order},\textit{po})=0$, and
$\#_{L_1}^{*}(\textit{Item},\textit{po})=1$.

\section{Object-Centric Behavioral Metrics}\label{sec:behavioral-metrics}
To introduce OCCGs, we first explain three types of object-centric behavioral metrics derived from an event log: \emph{ordering relation}, \emph{object involvement}, and \emph{performance} metrics.
Such metrics are used to define the semantics of OCCGs in \autoref{sec:pattern-graphs}.
% This section introduces three types of object-centric behavioral metrics derived from an event log: \emph{ordering relation}, \emph{object involvement}, and \emph{performance} metrics.
% The metrics are used to define the semantics of OCCGs introduced in \autoref{sec:pattern-graphs}.

% To this end, we first introduce behavioral properties of an OCEL: 1) \emph{ordering relations}, 2) \emph{object involvements}, and 3) \emph{diagnoses}.
% Using log characteristics, we define three ordering relations of two activities.
An ordering relation metric refers to the strength of a causal/concurrent/choice relation between two activities in an OCEL w.r.t. an object type.

\begin{definition}[Ordering Relation Metrics]\label{def:or}
Let $L$ be an object-centric event log.
For $ot \in \mi{types}(L)$ and $a,b\in \mi{acts}(L)$, we define the following ordering relation metrics of $L$:
\begin{compactitem}
    \item $ \mi{causal}_{\mi{L}}(ot,a,b)=
        \begin{cases}
            \frac{\#_{L}(ot,a,b)}{\#_{L}(ot,\{a,b\})}, \text{if }  \#_{L}(ot,\{a,b\}) > 0\\ 
            0, \text{otherwise}
        \end{cases} $
    \item $ \mi{concur}_{\mi{L}}(ot,a,b)=
        \begin{cases}
            1 - \frac{\mi{max}(\#_{L}(ot,a,b),\#_{L}(ot,b,a))-\mi{min}(\#_{L}(ot,a,b),\#_{L}(ot,b,a))}{\#_{L}(ot,a,b)+\#_{L}(ot,b,a)}, \\ \text{if }  \#_{L}(ot,a,b) + \#_{L}(ot,b,a) > 0 \\ 
            0, \text{otherwise}
        \end{cases} $
    \item $ \mi{choice}_{\mi{L}}(ot,a,b)=
        \begin{cases}
            1 {-} \frac{\#_{L}(ot,\{a,b\})+\#_{L}(ot,\{a,b\})}{\#_{L}(ot,\{a\})+\#_{L}(ot,\{b\})}, \text{if } \#_{L}(ot,\{a\}){+}\#_{L}(ot,\{b\}) {>} 0 \\ 
            0, \text{otherwise}
        \end{cases} $
\end{compactitem}
\end{definition}

$\mi{causal}_{\mi{L}}(ot,a,b)$, $\mi{concur}_{\mi{L}}(ot,a,b)$, and $\mi{choice}_{\mi{L}}(ot,a,b)$ all produce values between 0 (weak) and 1 (strong).
For $L_1$ in \autoref{tab:ocel}, $\mi{causal}_{\mi{L}_1}(\textit{Order},\textit{po},\textit{co})=1$, $\mi{concur}_{\mi{L}_1}(\textit{Order},\textit{po},\textit{co})=0$, $\mi{choice}_{\mi{L}_1}(\textit{Order},\textit{po},\textit{co})=0$, showing that \textit{po} and \textit{co} has a strong causal ordering relation.

Next, an object involvement metric quantitatively represents how the execution of an activity involves objects.

\begin{definition}[Object Involvement Metrics]\label{def:oi}
Let $L$ be an object-centric event log.
For $ot \in \mi{types}(L)$ and $a\in \mi{acts}(L)$, we define three object involvement metrics of $L$ in the following.
\begin{compactitem}
    \item $\mi{absent}_{\mi{L}}(ot,a)= \frac{\#_{L}^{0}(ot,a)}{|\mi{events}(a)|}$ is the strength of $ot$'s absence in $a$'s execution.
    % \item $\mi{present}_{\mi{L}}(ot,a)= 1-\frac{\#_{L}^{0}(ot,a)}{|\mi{events}(a)|}$ is the strength of type $ot$'s presence in the execution of $a$,
    \item $\mi{singular}_{\mi{L}}(ot,a)= \frac{\#_{L}^{1}(ot,a)}{|\mi{events}(a)|}$ is the strength of $ot$'s singularity in $a$'s execution.
    \item $\mi{multiple}_{\mi{L}}(ot,a)= \frac{\#_{L}^{*}(ot,a)}{|\mi{events}(a)|}$ is the strength of $ot$'s multiplicity in $a$'s execution.
\end{compactitem}
\end{definition}

All object involvement metrics produce values between 0 (weak) and 1 (strong).
For $L_1$ in \autoref{tab:ocel}, $\mi{absent}_{L_1}(\textit{Item},\textit{ec})=1$, showing that items are not involved in the execution of \textit{ec}. $\mi{singular}_{L_1}(\textit{Order},\textit{po})=1$ and $\mi{multiple}_{L_1}(\textit{Item},\textit{po})=1$, indicating that the execution of \textit{po} involves only one order and multiple items.

Finally, a performance metric refers to a performance/frequency value related to the execution of an activity.

\begin{definition}[Performance Metrics]\label{def:eoopn}
Let $L$ be an object-centric event log.
Let $\univ{\mi{measure}}$ be the universe of performance/frequency measure names, e.g., the average waiting time.
A performance metric of $\mi{L}$, $\mi{perf}_{\mi{L}} \in (\mi{acts}(L) \times \univ{measure}) \nrightarrow \mathbb{R}$, maps an activity and a performance/frequency measure to the value of the performance/frequency measure w.r.t. the activity.
\end{definition}

Note that we deliberately ``underspecify'' performance metrics, abstracting from the definition of individual performance metrics.
Performance metrics may include the average number of objects per object type for the execution of an activity (e.g., the average number of \textit{items} for placing an order), the average \emph{sojourn time} for the execution of an activity (e.g., the average sojourn time for confirming an order), etc.
For $L_1$ in \autoref{tab:ocel}, $\mi{perf}_{L_1}(\textit{po},\textit{avg-num-items})=3$, which denotes that the average number of \textit{items} for \textit{po} in $L_1$ is $3$.
Also, $\mi{perf}_{L_1}(\textit{co},\allowbreak \textit{avg-sojourn-time})=\textit{2 hours}$, which denotes that the average sojourn time for \textit{co} in $L_1$ is $2$ hours.
\section{Object-Centric Constraint Monitoring}\label{sec:pattern-graphs}
% primitives?
In this section, we explain our proposed approach to object-centric constraint monitoring.
To this end, we first introduce Object-Centric Constraint Graphs (OCCGs) to represent constraints.
% The semantics of OCCGs are defined with the object-centric behavioral metrics introduced in \autoref{sec:behavioral-metrics}.
Next, we introduce a monitoring engine to evaluate the violation of constraints represented by OCCGs by analyzing OCELs.

\subsection{Object-Centric Constraint Graphs (OCCGs)}
\begin{figure}[!htb]
    \centering
    \includegraphics[width=1\linewidth]{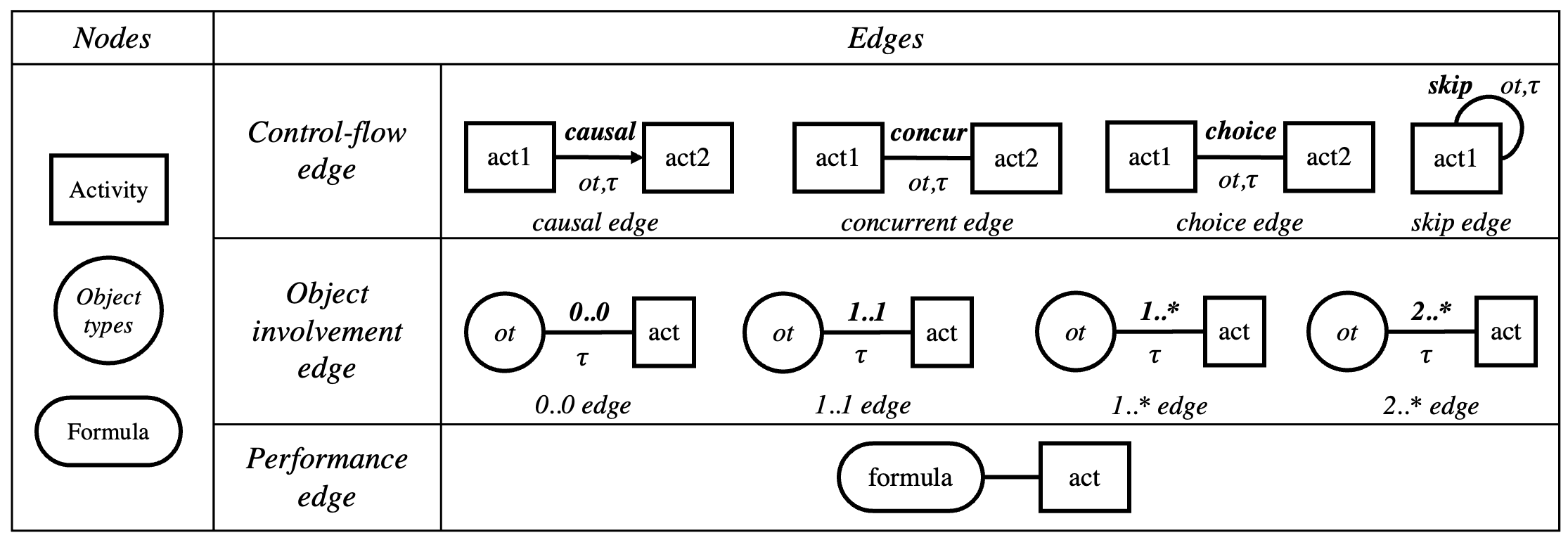}
    \caption{Graphical notations of OCCGs. $act \in \univ{act}$, $ot \in \univ{ot}$, and $\tau \in [0,1]$.}
    \label{fig:graphical_notation}
\end{figure}

An OCCG is a directed graph that consists of nodes and edges, as depicted in \autoref{fig:graphical_notation}.
Nodes consist of activities, object types, and \emph{formulas}.
A formula is a logical expression defined over performance measures of an activity using relational operators ($\le,\ge,=$) as well as logical operators such as conjunction ($\land$), disjunction ($\lor$), and negation ($\neg$).
Edges describe control-flow, object involvement, and performance edges.

\begin{definition}[Object-Centric Constraint Graph]
Let $F(X)$ be the set of all possible logical expressions with set $X$.
Let $A \subseteq \univ{act}$, $OT \subseteq \univ{ot}$, and $\mathcal{F} \subseteq F(\univ{measure})$.
Let $C=\{\mi{causal},\mi{concur},\mi{choice},\mi{skip}\}$ be the set of control-flow labels and $I=\{\mi{0..0}, \mi{1..1} \allowbreak , \mi{1..*}, \mi{2..*} \}$ the set of object involvement labels.
An object-centric constraint graph is a graph $\mi{cg} = (V,E_{\mi{flow}},E_{\mi{obj}},E_{\mi{perf}},l_{\mi{c}},l_{\mi{i}},l_{\tau})$ where
\begin{compactitem}
    \item $V \subseteq A \cup OT \cup \mathcal{F}$ is a set of nodes,
    \item $E_{\mi{flow}} \subseteq A \times OT \times A$ is a set of control-flow edges,
    \item $E_{\mi{obj}} \subseteq OT \times A$ is a set of object involvement edges,
    \item $E_{\mi{perf}} \subseteq \mathcal{F} \times A$ is a set of performance edges,
    \item $l_{\mi{c}} \in E_{\mi{flow}} \rightarrow C$ maps control-flow edges to control-flow labels such that, for any $(a,ot,b) \in E_{\mi{flow}}$, if $l_{\mi{c}}((a,ot,b))=\textit{skip}$, $a=b$,
    \item $l_{\mi{i}} \in E_{\mi{obj}} \rightarrow I$ maps object involvement edges to object involvement labels, and
    \item $l_{\tau} \in E_{\mi{flow}} \cup E_{\mi{obj}} \rightarrow [0,1]$ maps control-flow and object involvement edges to thresholds.
\end{compactitem}
$\univ{cg}$ denotes the set of all possible object-centric constraint graphs.
\end{definition}

\begin{figure}[!htb]
    \centering
    \includegraphics[width=1\linewidth]{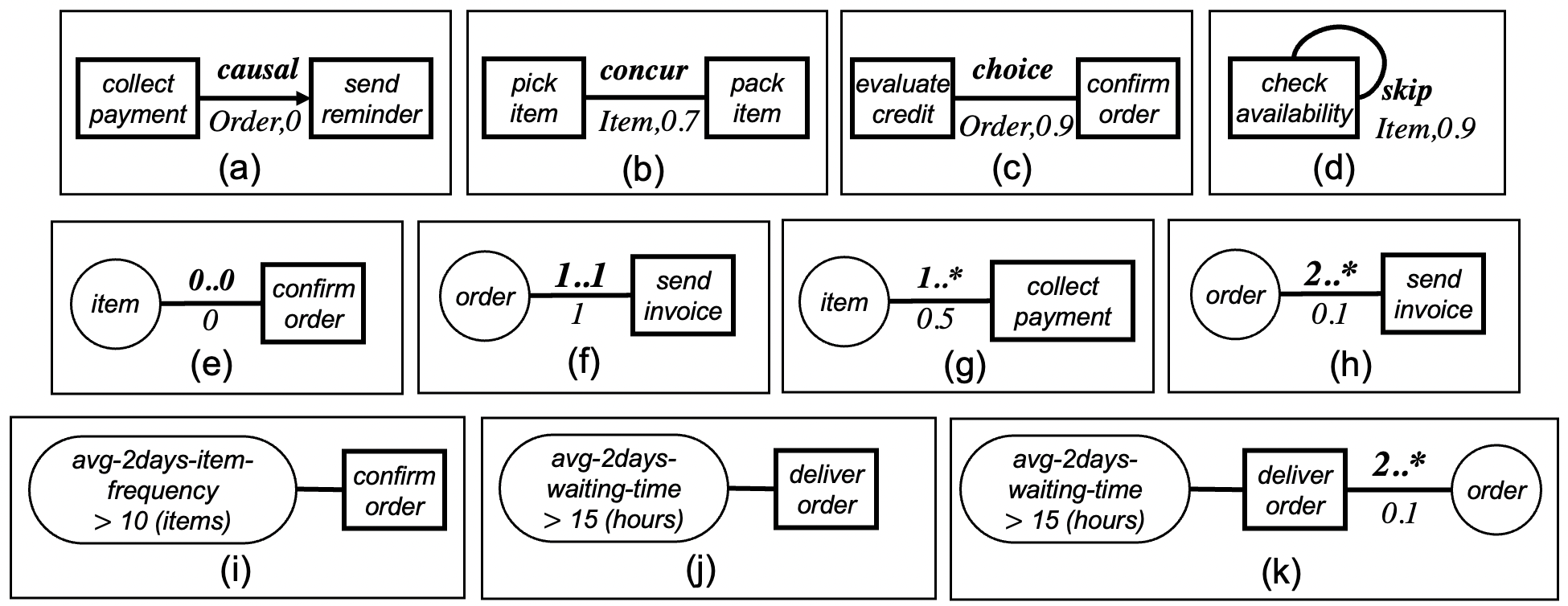}
    \caption{Examples of object-centric constraint graphs.}
    \label{fig:constraint-graph-examples}
\end{figure}

\autoref{fig:constraint-graph-examples}(a)-(k) introduces some example of OCCGs defined in an O2C process.
For instance, \autoref{fig:constraint-graph-examples}(a) is formally represented as follows: $\mi{cg}'=(V',E_{\mi{flow}}',\allowbreak \emptyset,\emptyset,l_{\mi{c}}',\emptyset,l_{\mi{\tau}}')$ where $V'=\{\textit{collect payment},\textit{send reminder}\}$, $E_{\mi{flow}}'=\{e_{1}=\allowbreak (\textit{collect } \allowbreak \textit{payment},\textit{Order},\allowbreak \textit{send reminder})\}$, $l_{\mi{c}}'( e_{1})=\textit{causal}$, and $l_{\tau}'( e_{1})=0$.

% \vspace{-0.5cm}
% \vspace{-0.5cm}

% \subsection{Semantics of Object-Centric Constraint Graphs}
We define the semantics of an OCCG with the notion of \emph{violation}.
% Using the behavioral properties of an OCEL, we define the semantics of an object-centric constraint graph with the notion of \emph{existence}.
An OCCG is \emph{violated} in an OCEL if nine conditions are satisfied, each of which corresponds to the different edges of the OCCG.

\begin{definition}[Semantics of object-centric constraint graphs]\label{def:semantics}
Let $L$ be an object-centric event log.
An object-centric constraint graph $\mi{cg} = (V,E_{\mi{flow}},E_{\mi{obj}}, \allowbreak E_{\mi{perf}},\allowbreak l_{\mi{c}},l_{\mi{i}},l_{\tau})$ is violated in $L$ if
\begin{compactenum}
    % \item $\forall_{e \in \{ e' \in E_{\mi{flow}} \mid \pi_{r}(e')=\mi{causal} \}} \ \mi{causal}_{\mi{L}}(ot,\pi_{s}(e),\pi_{t}(e)) \ge \tau$ where $(ot,\tau)=l_{\mi{flow}}(e)$,
    \item $\forall_{e=(a,ot,b) \in E_{\mi{flow}} \text{s.t. } ot {\in} \mi{types}(L) \land a{,}b {\in} \mi{acts}(L) \land l_{\mi{c}}(e)=\textit{causal}}  \ \mi{causal}_{\mi{L}}(ot,a,b) > l_{\tau}(e)$,
    \item $\forall_{e=(a,ot,b) \in E_{\mi{flow}} \text{s.t. } ot {\in} \mi{types}(L) \land a{,}b {\in} \mi{acts}(L) \land l_{\mi{c}}(e)=\textit{concur}}  \ \mi{concur}_{\mi{L}}(ot,a,b) > l_{\tau}(e)$,
    \item $\forall_{e=(a,ot,b) \in E_{\mi{flow}} \text{s.t. } ot {\in} \mi{types}(L) \land a{,}b {\in} \mi{acts}(L) \land l_{\mi{c}}(e)=\textit{choice}}  \ \mi{choice}_{\mi{L}}(ot,a,b) > l_{\tau}(e)$,
    \item $\forall_{e=(a,ot,a) \in E_{\mi{flow}} \text{s.t. } ot {\in} \mi{types}(L) \land a {\in} \mi{acts}(L) \land l_{\mi{c}}(e)=\textit{skip}}  \ 1 - \frac{\#_{L}(ot,\{a\})}{|\mi{objects}(ot)|} > l_{\tau}(e)$,
    \item $\forall_{e=(ot,a) \in E_{\mi{obj}} \text{s.t. } ot {\in} \mi{types}(L) \land a {\in} \mi{acts}(L) \land l_{\mi{i}}(e)=\mi{0..0}}
    \ \mi{absent}_{\mi{L}}(ot,a) > l_{\tau}(e)$,
    \item $\forall_{e=(ot,a) \in E_{\mi{obj}} \text{s.t. } ot {\in} \mi{types}(L) \land a {\in} \mi{acts}(L) \land l_{\mi{i}}(e)=\mi{1..1}}
    \ \mi{singular}_{\mi{L}}(ot,a) > l_{\tau}(e)$,
    \item $\forall_{e=(ot,a) \in E_{\mi{obj}} \text{s.t. } ot {\in} \mi{types}(L) \land a {\in} \mi{acts}(L) \land l_{\mi{i}}(e)=\mi{1..*}}
    \ 1-\mi{absent}_{\mi{L}}(ot,a) > l_{\tau}(e)$,
    \item $\forall_{e=(ot,a) \in E_{\mi{obj}} \text{s.t. } ot {\in} \mi{types}(L) \land a {\in} \mi{acts}(L) \land l_{\mi{i}}(e)=\mi{2..*}}
    \ \mi{multiple}_{\mi{L}}(ot,a) > l_{\tau}(e)$, and
    \item $\forall_{(f,a) \in E_{\mi{perf}} \text{s.t. } a {\in} \mi{acts}(L)} \ \textit{f evaluates to true w.r.t. } \mi{perf}_{\mi{L}}$.
\end{compactenum}
\end{definition}

For instance, \autoref{fig:constraint-graph-examples}(a) is violated if \textit{collect payment} is preceded by \textit{send reminder} at all w.r.t. \textit{Order}, \autoref{fig:constraint-graph-examples}(b) is violated if \textit{pick item} and \textit{pack item} are concurrently executed with the strength higher than $0.7$ w.r.t. \textit{Item}, 
% \autoref{fig:constraint-graph-examples}(c) is violated if \textit{evaluate credit} and \textit{confirm order} are exclusively executed with the strength higher than $0.9$ w.r.t. \textit{order},
\autoref{fig:constraint-graph-examples}(e) is violated if \textit{confirm order} is executed without involving \textit{Item} at all,
\autoref{fig:constraint-graph-examples}(k) is violated if the average waiting time of the last two days for \textit{deliver order} is longer than $15$ hours, and its execution involves multiple orders with the strength higher than $0.1$,
etc.

% Note that we can compose composite object-centric constraint graphs. 
% \autoref{fig:constraint-graph-examples}(k) shows a constraint graph with two conditions.
% The constraint is violated if both of the conditions are satisfied.
% It may indicate a problematic situation when the orders are duplicated while the bottleneck is developing in \textit{deliver order}.

\subsection{Monitoring Engine}
A monitoring engine analyzes the violation of OCCGs by analyzing an OCEL.

\begin{definition}[Monitoring Engine]
A monitoring engine $\mi{monitor} \in \univ{L} \times \univ{cg} \rightarrow \{\mi{true},\mi{false}\}$ is a function that maps an object-centric event log and an object-centric constraint graph to a Boolean value.
For any $L \in \univ{L}$ and $cg \in \univ{cg}$,  $\mi{monitor}(L,cg)=\mi{true}$ if $cg$ is $\textit{violated}$ in $\mi{L}$, and $\mi{false}$, otherwise.
\end{definition}

We implement the monitoring engine by 1) computing the object-centric behavioral metrics of an event log and 2) evaluating the violation of OCCGs based on them.
First, the derivation of ordering relation metrics and object involvement metrics is deterministic according to \autoref{def:or} and \autoref{def:oi}, respectively.
However, the computation of performance metrics is non-trivial. 
In this work, we use the approach proposed in~\cite{https://doi.org/10.48550/arxiv.2204.10662} to compute performance measures, such as \emph{sojourn time}, \emph{waiting time}, \emph{service time}, \emph{flow time}, \emph{synchronization time}, \emph{pooling time}, and \emph{lagging time}, and frequency measures, such as \emph{object count}, \emph{object type count}, etc.
% The approach 1) discovers an OCPN from an OCEL by following the approach presented in \cite{DBLP:journals/fuin/AalstB20}, 2) couples events in an OCEL to an OCPN by ``playing the token game'', and 3) computes performance measures such as \emph{flow time}, \emph{synchronization time}, \emph{pooling time}, and \emph{lagging time} along with more traditional metrics such as \emph{sojourn time}, \emph{waiting time}, \emph{service time}.
Finally, using the object-centric behavioral metrics, we evaluate the violation of OCCGs according to \autoref{def:semantics}.

\section{Implementation and Case Studies}\label{sec:evaluation}
This section presents the implementation of the approach presented in this paper and evaluates its feasibility by applying it to a production process and a P2P process of a real-life SAP ERP system.

\subsection{Implementation}
The approach presented in this work has been fully implemented as a web application\footnote{A demo video and sources: \url{https://github.com/gyunamister/ProPPa.git}} with a dedicated user interface.
The following functions are supported:

\begin{compactitem}
    \item Importing OCELs in different formats, including OCEL JSON, OCEL XML, and CSV.
    \item Designing object-centric constraint graphs using graphical tools.
    \item Computing object-centric behavioral metrics of OCELs and evaluating the violation of object-centric constraint graphs based on the metrics.
    \item Visualizing monitoring results with detailed analysis results.
\end{compactitem}

\subsection{Case Study: Production Process}
\begin{figure}[!htb]
    \centering
    \includegraphics[width=1\linewidth]{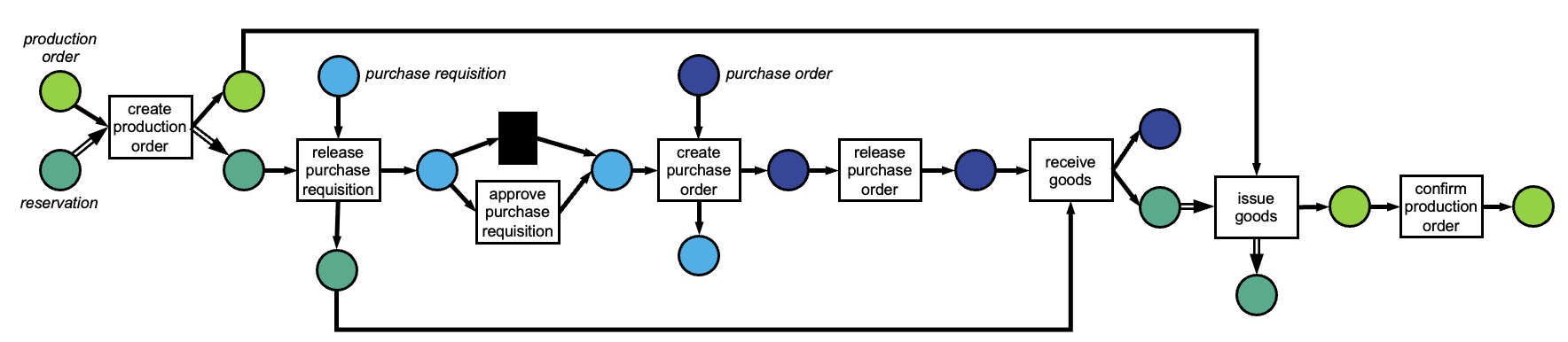}
    \caption{Production process: First, a production order is created with a variable number of reservations (i.e., required materials).
    Next, a purchase requisition is released and approved.
    Afterward, a purchase order is created based on the purchase requisition.
    Once the order is released, the reservations are received and issued for production. 
    Finally, the production order is confirmed.}
    \label{fig:production_ocpn}
\end{figure}

Using the implementation, we conduct a case study on a production process of a fictitious company supported by an SAP ERP system.
The process involves four object types: \textit{production order}, \textit{reservation}, \textit{purchase requisition}, and \textit{purchase order}.
\autoref{fig:production_ocpn} shows a process model of the production process using Object-Centric Petri Nets (OCPNs) as a formalism. We refer readers to \cite{DBLP:journals/fuin/AalstB20} for the details of OCPNs.

% \vspace{-0.5cm}
% \vspace{-0.5cm}

\begin{figure}[!htb]
    \centering
    \includegraphics[width=1\linewidth]{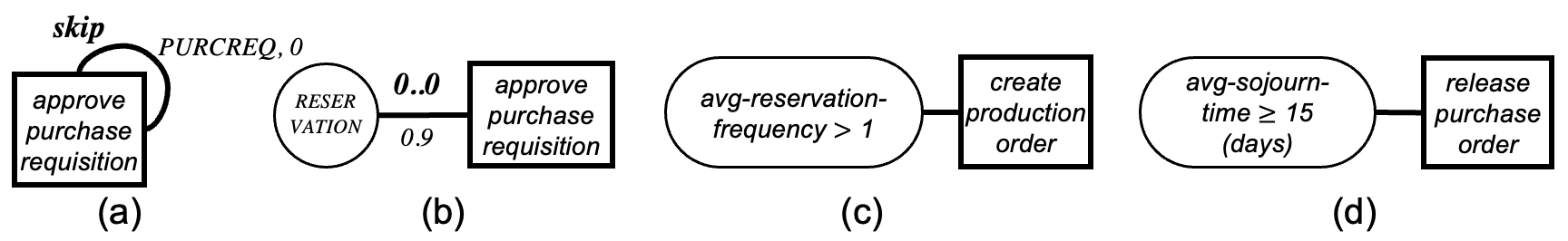}
    \caption{OCCGs representing the constraints of the production process.}
    \label{fig:eval_prod_patterns}
\end{figure}

We represent the following constraints using OCCGs:
\begin{compactitem}
    \item \textit{Skipping Purchase Requisition Approval (PRA)}; A purchase requisition should not skip the approval step at all.
    \autoref{fig:eval_prod_patterns}(a) represents the constraint.
    \item \textit{No reservation for Purchase Requisition Approval (PRA)}; The execution of \textit{approve purchase requisition} is supposed to include the corresponding reservation most of the time.
    \autoref{fig:eval_prod_patterns}(b) represents the constraint.
    \item \textit{Excessive reservations per Production Order (PO)}; The execution of \textit{create production order} should not involve more than one reservation on average. \autoref{fig:eval_prod_patterns}(c) represents the constraint.
    \item \textit{Delayed Purchase Order Release (POR)}; The average sojourn time of \textit{release purchase order} should be less than 15 days. \autoref{fig:eval_prod_patterns}(d) represents the constraint.
\end{compactitem}

We monitor the process using three OCELs extracted from the SAP ERP system.
Each event log contains events of different time windows; $L^{\mi{Jan22}}_{\mi{prod}}$, $L^{\mi{Feb22}}_{\mi{prod}}$, and $L^{\mi{Mar22}}_{\mi{prod}}$ contain events of Jan., Feb., and Mar. 2022\footnote{Event logs are publicly available at \url{https://github.com/gyunamister/ProPPa.git}.}.
\autoref{tab:eval-production} shows the monitoring result.
For instance, \textit{no reservation for PRA} and \textit{excessive reservations per PO} are violated for the three months. \textit{Skipping PRA} only is violated in the last two months, while \textit{delayed RPO} is violated only for Feb. 2022.

\begin{table}[!thb]
\centering
\caption{Monitoring results of the production process. \checkmark denotes the violation of the constraint in the corresponding event log.}
\label{tab:eval-production}
\resizebox{0.5\columnwidth}{!}{
\begin{tabular}{|c|ccc|}
\hline
\multirow{2}{*}{\textbf{Constraints}}     & \multicolumn{3}{c|}{\textbf{Event log}}                                                    \\ \cline{2-4} 
                                      & \multicolumn{1}{c|}{$\ L^{\mi{Jan22}}_{\mi{prod}} \ $} & \multicolumn{1}{c|}{$\ L^{\mi{Feb22}}_{\mi{prod}} \ $} & $\ L^{\mi{Mar22}}_{\mi{prod}} \ $ \\ \hline
\textit{Skipping PRA} & \multicolumn{1}{c|}{} & \multicolumn{1}{c|}{\checkmark}  &  \multicolumn{1}{c|}{\checkmark} \\ \hline
\textit{No reservation for PRA} & \multicolumn{1}{c|}{\checkmark}            & \multicolumn{1}{c|}{\checkmark}  & \multicolumn{1}{c|}{\checkmark}  \\ \hline
\textit{Excessive reservations per PO}   & \multicolumn{1}{c|}{\checkmark}  & \multicolumn{1}{c|}{\checkmark}  & \multicolumn{1}{c|}{\checkmark}  \\ \hline
\textit{Delayed POR} & \multicolumn{1}{c|}{} & \multicolumn{1}{c|}{\checkmark}  & \multicolumn{1}{c|}{}  \\ \hline
\end{tabular}
}
\end{table}

\subsection{Case Study: Procure-To-Pay (P2P) Process}
\begin{figure}[!htb]
    \centering
    \includegraphics[width=0.9\linewidth]{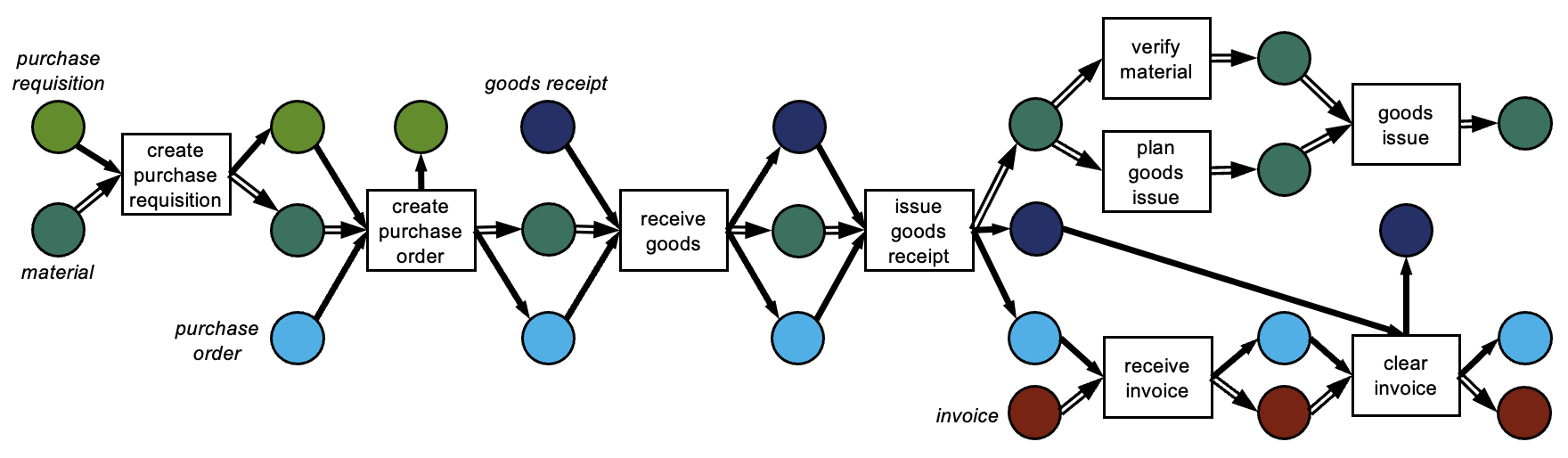}
    \caption{P2P process: First, a purchase requisition is created with multiple materials. 
    Next, a purchase order is created based on the purchase requisition and materials.
    Afterward, the materials are received and a goods receipt is issued.
    % A goods receipt is produced after receiving the materials of the purchase order.
    Then, the materials are verified and issued, and concurrently the invoice for the purchase order is received and cleared.}
    \label{fig:p2p_ocpn}
\end{figure}

Next, we explain a case study on the P2P process.
The process involves five object types: \textit{purchase requisition}, \textit{material}, \textit{purchase order}, \textit{goods receipt}, and \textit{invoice}.
\autoref{fig:p2p_ocpn} shows a process model of the process.

% \vspace{-0.5cm}
% \vspace{-0.5cm}

\begin{figure}[!htb]
    \centering
    \includegraphics[width=1\linewidth]{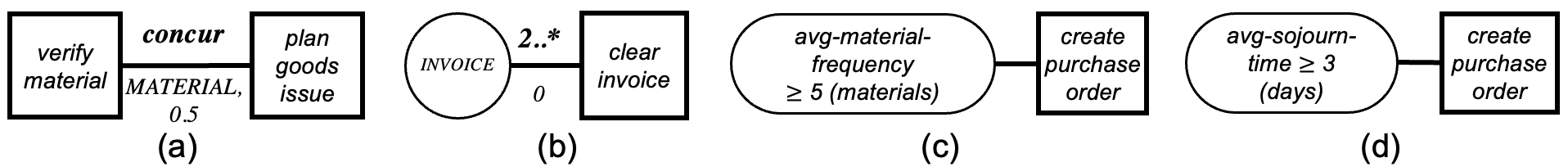}
    \caption{OCCGs representing the constraints of the P2P process.}
    \label{fig:eval_p2p_patterns}
\end{figure}

We represent the following constraints using OCCGs:
\begin{compactitem}
    \item \textit{Concurrency between Verify Material (VM) and Plan Goods Issue (PGI)}; VM and PGI are usually not supposed to be concurrently executed.
    \autoref{fig:eval_p2p_patterns}(a) represents the constraint.
    \item \textit{Clearance of multiple invoices}; The execution of \textit{clear invoice} should not involve multiple invoices at all.
    \autoref{fig:eval_p2p_patterns}(b) represents the constraint.
    \item \textit{Excessive materials per Purchase Order (PO)}; The execution of \textit{create purchase order} should involve less than five materials on average.
    \autoref{fig:eval_p2p_patterns}(c) represents the constraint.
    \item \textit{Delayed Purchase Order Creation (POC)}; The average sojourn time of \textit{create purchase order} should be less than three days.
    \autoref{fig:eval_p2p_patterns}(d) represents the constraint.
\end{compactitem}

We monitor the process using three OCELs extracted from the SAP ERP system.
Each event log contains events of different time windows; $L^1_{\mi{p2p}}$ starting from \text{\scriptsize 01-Aug-2021} and ending at \text{\scriptsize 14-Oct-2021}, $L^2_{\mi{p2p}}$ starting from \text{\scriptsize 15-Oct-2021} and ending at \text{\scriptsize 18-Jan-2022}, and $L^3_{\mi{p2p}}$ starting from \text{\scriptsize 01-Feb-2022} and ending at \text{\scriptsize 16-May-2022}.
\autoref{tab:eval-p2p} shows the monitoring result.
\textit{Concurrency between VM and PGI} and \textit{clearance of multiple invoices} are only violated in the first two time windows, whereas \textit{Excessive materials per PO} and \textit{delayed POC} are only violated in the last time window.

% \vspace{-1cm}
\begin{table}[!thb]
\centering
\caption{Monitoring results of the P2P process. \checkmark denotes the violation of the constraint in the event log.}
\label{tab:eval-p2p}
\resizebox{0.5\columnwidth}{!}{
\begin{tabular}{|c|ccc|}
\hline
\multirow{2}{*}{\textbf{Constraints}}     & \multicolumn{3}{c|}{\textbf{Event log}}                                                    \\ \cline{2-4} 
                                      & \multicolumn{1}{c|}{$\ L^1_{\mi{p2p}} \ $} & \multicolumn{1}{c|}{$\ L^2_{\mi{p2p}} \ $} & $\ L^3_{\mi{p2p}} \ $ \\ \hline
\textit{Concurrency between VM and PGI} & \multicolumn{1}{c|}{\checkmark}  & \multicolumn{1}{c|}{\checkmark}  &  \multicolumn{1}{c|}{} \\ \hline
\textit{Clearance of multiple invoices} & \multicolumn{1}{c|}{\checkmark} & \multicolumn{1}{c|}{\checkmark}  & \multicolumn{1}{c|}{}  \\ \hline
\textit{Excessive materials per PO}   & \multicolumn{1}{c|}{}  & \multicolumn{1}{c|}{}  &  \multicolumn{1}{c|}{\checkmark} \\ \hline
\textit{Delayed POC} & \multicolumn{1}{c|}{} & \multicolumn{1}{c|}{}  & \checkmark  \\ \hline
\end{tabular}
}
\end{table}
% \vspace{-0.5cm}
\section{Conclusion}\label{sec:conclusion}
In this paper, we proposed an approach to process-level object-centric constraint monitoring.
To this end, we first introduced object-centric behavioral metrics and defined OCCGs using the metrics.
The proposed monitoring engine evaluates the constraints represented by OCCGs by analyzing OCELs.
We have implemented the approach as a Web application and discussed two case studies.

This paper has several limitations.
The suggested object-centric constraint graphs only represent the constraints selectively introduced in this work.
More advanced constraints are not considered, e.g., ordering relations with the temporality (e.g., eventual or direct causality).
Also, constraint graphs do not support timed constraints, e.g., no involvement of an object type during a specific period of time.
In future work, we plan to extend the proposed approach by including more advanced constraints.
Another interesting direction of future work is to apply the proposed approach to real-life business processes.
\vspace{-0.2cm}

\paragraph{\textbf{Acknowledgements}}
We thank the Alexander von Humboldt (AvH) Stiftung for supporting our research.

\bibliographystyle{splncs04}
\bibliography{mybib}

\end{document}